\documentclass[conference]{IEEEtran}
\IEEEoverridecommandlockouts
\usepackage{cite}
\usepackage{amsmath,amssymb,amsfonts}
\usepackage{algorithm}
\usepackage{algorithmic}
\usepackage{graphicx}
\usepackage{textcomp}
\usepackage{xcolor}
\DeclareMathOperator{\sinc}{sinc}
\def\BibTeX{{\rm B\kern-.05em{\sc i\kern-.025em b}\kern-.08em
    T\kern-.1667em\lower.7ex\hbox{E}\kern-.125emX}}
\begin{document}

\title{Development of a Simulation Environment for Evaluation of a Forward Looking Sonar System for Small AUVs}

\author{\IEEEauthorblockA{Christopher Morency, Daniel J. Stilwell}
\IEEEauthorblockA{Bradley Department of Electrical and Computer Engineering \\
Virginia Polytechnic Institute and State University\\
Blacksburg, VA, USA \\
\{cmorency, stilwell\}@vt.edu \and Sebastian Hess \\ Atlas Elektronik GmbH \\ Bremen, Germany \\ sebastian.hess@atlas-elektronik.com}
}

\maketitle

\begin{abstract}
    This paper describes a high-fidelity sonar model and a simulation environment that implements the model. The model and simulation environment have been developed to aid in the design of a forward looking sonar for autonomous underwater vehicles (AUVs). The simulator achieves real-time visualization through ray tracing and approximation. The simulator facilitates the assessment of sonar design choices, such as beam pattern and beam location, and assessment of obstacle detection and tracking algorithms. An obstacle detection model is proposed for which the null hypothesis is estimated from the environmental model. Sonar data is generated from the simulator and compared to the expected results from the detection model demonstrating the benefits and limitations of the proposed approach. 
\end{abstract}

\section{Introduction}

We describe a high-fidelity sonar model that is well-suited to evaluation of design choices for forward-looking sonar. We present a complete set of equations that constitute the model as well as approaches for implementing the model in a numerical simulation. We are especially interested in assessing the performance of forward-looking sonar systems that could be used for object detection in small autonomous underwater vehicle (AUV) applications. Our simulation combines a high-fidelity sonar model with the capability to simulate AUV missions in a three dimensional environment in real time. 

Many open source and commercial high-fidelity sonar models and simulations are described in the literature \cite{APL-UW}, \cite{Espresso}, \cite{LYBIN}. The Sonar Simulation Toolset (SST) \cite{APL-UW} from APL-UW is a high-fidelity open-source sonar simulation toolset. The SST simulates an ocean environment and the sound generated by a sonar. Due to the complexity of the model, it is not intended for use as a real time software. Espresso \cite{Espresso} was built to evaluate NATO minehunting sonar performance. Espresso makes a flat and homogeneous seabed assumption and does not include sonar motion in the model \cite{Couillard_2014}. LYBIN \cite{LYBIN} is an acoustic ray-theoretical model for sonar performance. LYBIN uses a high-fidelity sonar model and runs in real time, however, it is limited to two-dimensions. These simulations provide the desired high-fidelity sonar models and excel at modelling sonar performance yet are not capable of evaluating sonar performance through simulated AUV missions in real time. 

AUV simulators \cite{Prats_2012}, \cite{Manhaes_2016}, \cite{Braunl_2004} make up a separate class of simulators, mainly focused on rapid protyping of AUVs by simulating the dynamics and missions for AUVs. UWSim \cite{Prats_2012} is a visualization and simulation tool which uses a range camera to simulate sonar data, capturing the distance of objects from the sensor. UUV Simulator \cite{Manhaes_2016} is a Gazebo-based package for AUV simulation. The Gazebo-based simulation provides a sensor model for a multi-beam echo sounders using 2D laser range finders, returning the distance of an object from the sensor. Project Mako \cite{Braunl_2004} describes an AUV simulator (SubSim) built for an AUV competition. The SubSim sensor model traces a ray from the sonar, only returning the distance to the object. These simulators compute the distance to an object, but they do not calculate the sound intensity returned to the sonar. Therefore, they are unsuitable for assessing obstacle detection/tracking algorithms.

To assess the performance of object detection algorithms, we seek a high-fidelity forward-looking sonar simulator that can be integrated with an accurate AUV motion model. Our sonar model calculates the sound intensity received by each sonar transducer element, binned by distance, for the entire range of each sonar ping. Our numerical simulation can be used to test the various types of design choices, such as the number and direction of beams for obstacle detection and tracking. 

Obstacle detection algorithms can be assessed using the high-fidelity simulated sonar model. We briefly illustrate the assessment of an obstacle detection approach based on the Bayesian framework and employ a Bayesian detector to construct decision rules. When all uncertain parameters are known, our Bayesian detector is optimal. A similar framework is presented in \cite{c22}.

We consider a forward looking sonar with a limited number of beams as a case study throughout this paper. This illustrative example falls between forward looking imaging sonars, such as the  Blueview P450-15E \cite{Horner_2009} and DIDSON \cite{Belcher_2002} sonars which use a large number of fixed beams, and a single beam forward looking sonar that is stationary \cite{Hutin_2005} or mechanically steered such as the Imagenex 881L Profiling Sonar \cite{Heidarsson_2011}. We assess the accuracy of our numerical simulation by comparing the results from the simulation to the theoretical results obtained through our environmental model, which cannot be run in real-time. Using a limited number of beams, the simulator provides a real-time visualization capability. The number of beams used in the simulator can be scaled up at the cost of additional computational effort. 

Organization of the paper is as follows. The equations used to model sound propagation appear in Section \ref{EM}. The detection model outlines approaches to compute the null and alternate hypothesis in Section \ref{DM}. Discussion of the simulator and results from illustrative test cases are presented in Sections \ref{SIM} and \ref{RES}.


\section{Environmental model}\label{EM}

The environmental model is constructed such that the sound energy returned to the sonar from reflections can be characterized by a discrete set of distances, or equivalently, discrete times. The energy returned to the sonar is discretized into a set of equal length distance bins over the entire range of the sonar. The environmental model consists of a set of equations that model the acoustic propagation of the sonar. These equations model the sound velocity, transmission loss, beam pattern loss, backscatter, sonar resolution and noise. The transmission loss is comprised of attenuation and spread loss. The backscatter is the energy reflected back to the sonar from bottom, surface and volume. 

\subsection{Sound Velocity}

The speed of sound can be estimated with less than 0.1 m/s error using the empirical formals in \cite{Chen_and_Millero_1977} and \cite{Del_Grosso_1974}.  However, the empirical formulas are difficult to compute in real-time, so we adopt a simplified approximation for the speed of sound, described in \cite{Medwin_1975}. Sound velocity in m/s is expressed

\begin{equation}\label{SOS}
    \begin{split}
        c = 1449.2 + 4.6T - 0.055T^2 + 0.00029T^3 \\+ (1.34 - 0.010T)(S - 35) + 0.016z
    \end{split}
\end{equation}
where $T$ is temperature (\textdegree C), $S$ is salinity (ppt), and $z$ is water depth (m). The equation \eqref{SOS} is valid for $0\leq T \leq 35$, $0$ ppt $\leq S \leq 45$ ppt, and $0$m $\leq z \leq$ $1000$m.

\subsection{Attenuation}

Attenuation is modeled using the Francois and Garrison formulas \cite{z1}, \cite{z2},

\begin{equation*}
    \alpha_w = \frac{A_1P_1f_1f^2}{f_1^2 + f^2} + \frac{A_2P_2f_2f^2}{f_2^2 + f^2} + A_3P_3f^2\text{ dB/km}
\end{equation*}
The boric acid coeficients are

\begin{equation*}
    A_1 = \frac{8.696}{c}10^{0.78pH-5}
\end{equation*}

\begin{equation*}
    f_1 = 2.8\sqrt{\frac{S}{35}}10^{4-\frac{1245}{T+273}}
\end{equation*}

\begin{equation*}
    P_1 = 1
\end{equation*}
The magnesium sulphate coefficients are

\begin{equation*}
    A_2 = 21.44\frac{S}{c}(1+0.025T)
\end{equation*}

\begin{equation*}
    f_2 = \frac{8.17 \times 10^{8-1990/(T+273)}}{1+0.0018(S-35)} 
\end{equation*}

\begin{equation*}
    P_2 = 1-1.37\times 10^{-4}z_{\max}+6.2\times 10^{-9}\times z_{\max}^2
\end{equation*}
The coefficients for pure water viscocity are

\begin{equation*}
    A_3 = \begin{cases} 
    4.937\times 10^{-4} - 2.59\times 10^{-5}T + \\ 9.11 \times 10^{-7}T^2 - 1.5 \times 10^{-8}T^3 & \text{for }T \leq 20 \text{\textdegree C}
    \\
    3.964\times 10^{-4} - 1.146\times 10^{-5}T + \\ 1.45 \times 10^{-7}T^2 - 6.5 \times 10^{-10}T^3 & \text{for }T > 20 \text{\textdegree C} 
    \end{cases}
\end{equation*}

\begin{equation*}
    P_3 = 1-3.83\times 10^{-5}z_{\max}+4.9\times 10^{-10}\times z_{\max}^2
\end{equation*}
where $f$ is the frequency in kHz, $T$ is the temperature (\textdegree C), $S$ is salinity (ppt), $z_{\max}$ is the maximum water depth (m), $c$ is the sound speed (m/s) and $pH$ is the acidity (Moles/litre). The total attenuation $\alpha_{t}$ with respect to distance $d$ (m) is

\begin{equation}\label{attenuation}
    \alpha_{t}= \frac{(2d-1)\times \alpha_w}{1000} \text{ dB}
\end{equation}

\subsection{Spread Loss}

Assuming spherical spreading with no cylindrical spreading, the intensity of a sound wave is inversely proportional to the distance $d$ (m) \cite{Jenson_2011}. Since the sound energy in an area is computed with respect to the energy at $1$ meter, the two-way loss due to spherical spreading for distance $d$ is

\begin{equation}\label{spreadloss}
S_L = 40\log_{10}(d)
\end{equation}
The total two-way transmission loss is

\begin{equation}\label{TL}
    TL = S_L + \alpha_t
\end{equation}
where $\alpha_t$ is the attenuation at distance $d$ from \eqref{attenuation}. 

\subsection{Beam Pattern}

The beam pattern is calculated using the single-point-source (SPS) approach \cite{Marage_2010}. The calculation takes into consideration the wavelength $\lambda$ (m), the transducer's horizontal length $L_H$ (m), the transducer's vertical length $L_V$ (m), the horizontal beam angle $\theta$ (radians), and vertical beam angle $\psi$ (radians). The beam pattern loss is 

\begin{equation*}
    BP = 20\log_{10}(\alpha \beta) \text{ dB}
\end{equation*}

\noindent where

\begin{equation}\label{bp2}
    \alpha = \sinc \left(\sin(\theta)\cos(\psi)\frac{L_H}{\lambda}\right)
\end{equation}

\noindent and 

\begin{equation}\label{bp3}
    \beta = \sinc \left(\sin(\psi) \frac{L_V}{\lambda} \right)
\end{equation}
The wave-length $\lambda = \frac{c}{f}$ is a dependent on the speed of sound in water $c$ (m/s) and the frequency $f$ (kHz).

\subsection{Bottom Backscatter}\label{BB EM}

Models of bottom backscattering are used to estimate the intensity of the sonar signal that is reflected back to the sonar from the sea floor \cite{Urick_1983}. Bottom backscatter is dependent on the bottom type $bt$, grazing angle $\Theta$ (radians) and frequency $f$ (kHz). Reverberation is the total reradiated acoustic energy caused by inhomogeneities in the ocean. The bottom reverberation received by the sonar transducer element is

\begin{equation}\label{bb1}
    RL_B = SL - TL + BP_T + BP_R + RS_B
\end{equation}

\noindent where $SL$ is the source level, $BP_T$ is the average loss from the beam pattern of the transmitter, $BP_R$ is the average loss from the beam pattern of the receiver, $TL$ is the two-way transmission loss, and $RS_B$ is the reverberation strength of the bottom (all in dB). The reverberation strength $RS_B$ can be computed

\begin{equation}\label{bb2}
    RS_B = S_B + 10\log_{10}(A_B)
\end{equation}

\noindent where $A_B$ denotes the ensonified area of the bottom ($\text{m}^2$). The bottom backscatter coefficient $S_B$ is found using the SEARAY model \cite{Kraus}, for which the reverberation coefficient is defined 

\begin{equation*}
    S_B = 10\log_{10} (3.03\beta f^{3.2-0.8bt}10^{2.8bt-12}+ 10^{-4.42}) \text{ dB}/\text{m}^2
\end{equation*}

\noindent where

\begin{equation*}
    \beta = \gamma(\sin(\Theta)+0.19)^{(bt)\cos^{16}(\Theta)}
\end{equation*}

\noindent and 

\begin{equation*}
    \gamma = 1+ 125e^{-2.64(bt-1.75)^2 -\frac{50}{bt}\cot^2(\Theta)}
\end{equation*}

\noindent Common bottom type values \cite{Kraus} are

\begin{equation*}
    bt =\begin{cases}
    1 & \text{mud} \\
    2 & \text{sand} \\
    3 & \text{gravel} \\
    4 & \text{rock}
    \end{cases}
\end{equation*}

\subsection{Surface Backscatter}

Surface backscattering defines how much of the signal is reflected back to the sonar from the sea surface \cite{Urick_1983}. Surface backscatter is dependant on the grazing angle $\Theta$ (radians), frequency $f$ (kHz) and wind speed $v_w$ (knots). The total surface reverberation received by the sonar transducer element is

\begin{equation}\label{sb1}
    RL_S = SL + BP_T + BP_R - TL + RS_S
\end{equation}

\noindent where $SL$ is the source level, $BP_T$ is the average loss from the beam pattern of the transmitter, $BP_R$ is the average loss from the beam pattern of the receiver, $TL$ is the two-way transmission loss, and $RS_S$ is the reverberation strength of the surface (all in dB). Reverberation strength due to the surface is 

\begin{equation}\label{sb2}
    RS_S = S_S + 10\log_{10}(A_S)
\end{equation}
where $A$ denotes the total ensonified area of the surface ($\text{m}^2$). The surface backscattering coefficient \cite{Kraus} is 

\begin{equation*}
    \begin{split}
            &S_S = \\ &10\log_{10}\left(10^{-5.05}(1+v_w)^2(f+0.1)^{v_w/150}\tan^\beta(\Theta)\right) \text{ dB/$\text{m}^2$}
    \end{split}
\end{equation*}
where

\begin{equation*}
    \beta = 4 \left( \frac{v_w + 2}{v_w + 1} \right) + \left( 2.5(f+0.1)^{-1/3} - 4\right)\cos^{1/8}(\Theta)
\end{equation*}

\subsection{Volume Backscatter}

Volume backscattering is the intensity of the signal that is reflected back towards the sonar through a volume of water \cite{Urick_1983}. This phenomenon arises from biological organisms and turbidity. It is dependant on the frequency $f$ (kHz) and particle density $Sp$ (dB). The total contribution from the volume reverberation received by the sonar transducer element is

\begin{equation}\label{vb1}
    RL_V = SL - TL + BP_T + BP_R + RS_V \text{ dB} 
\end{equation}

\noindent where $SL$ is the source level, $BP_T$ is the average loss from the beam pattern of the transmitter, $BP_R$ is the average loss from the beam pattern of the receiver, $TL$ is the two-way transmission loss and $RS_V$ is the reverberation strength of the volume (all in dB). $RS_V$ is

\begin{equation}\label{vb2}
    RS_V = S_V +10\log_{10}(V) \text{ dB} 
\end{equation}

\noindent where $V$ is the total ensonified volume. The volume reverberation coefficient \cite{Kraus} is

\begin{equation*}
    S_V = Sp +7\log_{10}(f) \text{ dB}/\text{m}^3 
\end{equation*}
where 

\begin{equation*}
    Sp = \begin{cases}
    -50dB & \text{High particle density} \\
    -70dB & \text{Moderate particle density} \\
    -90dB & \text{Low particle density} \\
    \end{cases}
\end{equation*}

\subsection{Sonar Resolution}

The duration of a sonar ping is proportional to the distance the wave front covers at any given time \cite{Medwin_1998}. This phenomenon determines the size of the area for which the energy reflected back to the sonar may have originated at a given time. Using CHIRP pulses, the sonar resolution $\delta_y$ (m) is computed

\begin{equation}\label{resolution}
    \delta_y = \frac{c}{2B}
\end{equation}
where $c$ (m/s) is speed of sound in water and $B$ (Hz) is the bandwidth. 

\subsection{Noise}

Isotropic noise power has numerous contributions \cite{Brekhovkikh_2003} which have been examined theoretically and compared to experimental data \cite{Wenz_1962}, \cite{Mellen_1952}. The noise sources considered are described by the following empirical formulas \cite{Coates_1990} and are turbulence noise, 

\begin{equation*}
    NL_{\text{turb}} = 17 - 30\log_{10}(f) \text{ dB}
\end{equation*}
shipping traffic noise,

\begin{equation*}
\begin{split}
        &NL_{\text{traffic}} = \\&40 + 20(D - 0.5) + 26\log_{10}(f) - 60\log_{10}\left( f+0.03\right) \text{ dB}
\end{split}
\end{equation*}
sea state noise,

\begin{equation*}
    NL_{\text{ss}} = 50 +5.38w^{0.5} + 20\log_{10}(f) -40\log_{10}(f+0.4) \text{ dB}
\end{equation*}
and thermal noise,

\begin{equation*}
    NL_{\text{therm}} = -15 + 20\log_{10}(f) \text{ dB}
\end{equation*}
where $v_W$ is the wind speed (knots), $f$ is the frequency (kHz), and $D$ is the shipping density between 0 (very light) and 1 (heavy). The isotropic noise level $NL$ around a 1 Hz frequency band is expressed

\begin{equation*}
\begin{split}
    &NL = \\&10\log_{10} \left(10^{\frac{NL_{\text{turb}}}{10}} + 10^{\frac{NL_{\text{traffic}}}{10}} + 10^{\frac{NL_{\text{ss}}}{10}} + 10^{\frac{NL_{\text{therm}}}{10}}\right)
\end{split}
\end{equation*}
The noise level over a frequency band $B$ (Hz) is 

\begin{equation*}
    NL_B = NL + 10\log_{10}(B)
\end{equation*}
The contributions from rainfall noise, biological noise and vessel noise from the sonar platform are not taken into consideration in our model.

\section{Detection Model}\label{DM}

A detection model is constructed to identify potential objects in the field of view of the AUV. The model maps the measurements from the sonar to the probability of an obstacle being present. Bayes theorem can be used as the basis of obstacle detection. Given a measurement, $z_t$, the probability of an obstacle at a certain location $s_i$ is  
\begin{equation*}
    p(s_i=1|z_t) = \frac{p(z_t|s_i=1)p(s_i=1)}{\sum_{X \in \{0,1\}} p(z_t|s_i=X)p(s_i=X)}
\end{equation*}
where $p(s_i = 1)$ is the prior probability of an obstacle being present. 

Since the probability of an obstacle in the field of view of the AUV is unknown \textit{a priori}, a likelihood ratio

\begin{equation*}
    \lambda_d = \frac{p(z_t|s_i=1)}{p(z_t|s_i=0)}
\end{equation*}
is used to compare the ratio of the null and the alternative hypothesis. The detection threshold $\gamma_d$ is selected for the desired sensitivity of the system. The sensitivity can vary depending on the mission since, as the sensitivity increases, the probability of false alarm increases as well. The probability of detection and false alarm are

\begin{equation*}
    P_D = \int_{\gamma_d}^\infty p(z_t|s_i=1) dz_t
\end{equation*}

\begin{equation*}
    P_{FA} = \int_{\gamma_d}^\infty p(z_t|s_i=0) dz_t
\end{equation*}
and the detection algorithm selects

\begin{equation*}
    s_i = 
    \begin{cases}
    1 & \lambda_d \geq \gamma_d \\
    0 & \text{otherwise}
    \end{cases}
\end{equation*}

\subsection{Computation of the Null Hypothesis}
A set of equations to compute the null hypothesis are presented and applied in Section \ref{RES} to evaluate the performance of the simulator. When no obstacle is present, the expected return from a ping has three contributions: bottom backscatter, surface backscatter and volume backscatter. The expected intensity given the null is

\begin{equation}\label{expected_returns}
\mathbb{E}\left[z_t|s_i = 0\right] = 10\log_{10}\left(10^{\frac{RL_B}{10}} + 10^{\frac{RL_S}{10}} + 10^{\frac{RL_V}{10}} \right)
\end{equation}
where $RL_B$, $RL_S$, and $RL_V$ are the contributions from \eqref{bb1}, \eqref{sb1}, and \eqref{vb1} respectively. 

\subsection{Bottom Backscatter Area Calculation}\label{s10}

\begin{figure}[htbp]
\centering 
\includegraphics[width=3.0in]{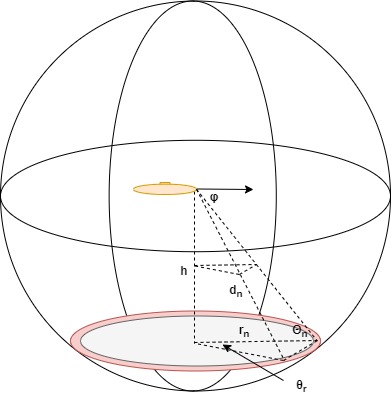}
\caption{Bottom Backscatter Diagram}
\label{fig}
\end{figure}

The total contribution from bottom backscattering is expressed in \eqref{bb1}. From \eqref{bb2}, the total ensonified area for each distance bin is calculated, where the distance bin is the discretization of the data from the sonar. We assume that height can be measured and the bottom is locally flat.  

Directly underneath the vehicle, we presume a circular ensonified area for the distance bin $n$ containing the current height $h$ (m). The radius of the ensonified area for distance bin $n$ is

\begin{equation*}
    r_n = \begin{cases} \sqrt{d_n^2 - h^2} &  h < d_n
                        \\ 0 & h \geq d_n
    \end{cases}
\end{equation*}
where $d_n$ is the distance to the end of distance bin $n$. Using the radius, the area of distance bin $n$ is

\begin{equation}\label{bb_A}
    A_{n} = \begin{cases} \pi r_{n}^2 - A_{n-1} &  h < d_n
                        \\ 0 & h \geq d_n
    \end{cases}
\end{equation}
where $A_0 = 0$. The grazing angle $\Theta$ to the center of the ring for distance bin $n$ is

\begin{equation*}
    \Theta_{n} =\begin{cases} \sin^{-1} \left(\frac{2|h|}{d_{n}+d_{n-1}}\right) &  h < d_n \\ 0 & h \geq d_n
    \end{cases}
\end{equation*}

An additional signal loss occurs due to the physical properties of the transducer, referred to as the beam pattern loss. To calculate the effect of the beam pattern, the horizontal beam pattern angle $\theta$, and the vertical beam pattern angle $\psi$ are required. A coordinate system is attached to the sonar transducer element such that the vector $\begin{bmatrix}1 & 0 & 0\end{bmatrix}^{\mathsf{T}}$, expressed in the coordinate frame, points directly away from the face of the transducer element. The vector from the sonar to each point along the circle is 
\begin{equation*}
    v_b = \begin{bmatrix} 
    \left(\frac{r_{n} + r_{n-1}}{2}\right)\cos(\theta_r) \\ \left(\frac{r_{n} + r_{n-1}}{2}\right)\sin(\theta_r) \\ h
    \end{bmatrix}
\end{equation*}
where $-\pi \leq \theta_r \leq \pi$.  We use a rotation matrix to rotate each beam vector along the ring to the sonar frame. For example, a sonar directed $\theta_p$ degrees downwards from $\begin{bmatrix}1 & 0 & 0\end{bmatrix}^\mathsf{T}$ is computed
\begin{equation*}
    v = \begin{bmatrix}
    \cos(\theta_p) & 0 & \sin(\theta_p) 
    \\ 0 & 1 & 0 
    \\-\sin(\theta_p) & 0 & \cos(\theta_p)
    \end{bmatrix}
    \begin{bmatrix} 
    \left(\frac{r_{n} + r_{n-1}}{2}\right)\cos(\theta_r) \\ \left(\frac{r_{n} + r_{n-1}}{2}\right)\sin(\theta_r) \\ h
    \end{bmatrix}
\end{equation*}
where the entries of $v$ are denoted $v = \begin{bmatrix}v_x & v_y & v_z\end{bmatrix}^\mathsf{T}$. The horizontal beam pattern angle $\theta$ is 
\begin{equation*}
    \theta =  \tan^{-1} \left(\frac{v_y}{v_x}\right)
\end{equation*}
The vertical beam pattern angle $\psi$ is 
\begin{equation*}
    \psi =  \tan^{-1} \left(\frac{v_z}{\sqrt{v_x^2+v_y^2}}\right)
\end{equation*}
The average loss from the beam pattern around the ring is given by 

\begin{equation*}
    \frac{1}{\pi}\int_{-\pi}^{\pi}BP(\theta, \psi )d\theta_r
\end{equation*}
where the beam pattern is defined 

\begin{equation}\label{BP_bbh}
    BP(\theta, \psi) = 
    \begin{cases}
    20\log_{10}(\alpha \beta) \text{ dB} &  -\frac{\pi}{2} < \theta, \psi < \frac{\pi}{2}\\
    0 & \text{otherwise}
    \end{cases}
\end{equation}
The variables $\alpha$ and $\beta$ are \eqref{bp2} and \eqref{bp3}.

To compute the contribution from bottom backscattering \eqref{bb1} for each distance bin, $BP_T$ and $BP_R$ are computed with \eqref{BP_bbh} and the ensonified area $A_B$ \eqref{bb2} is $A_n$ \eqref{bb_A}. The distance $d$ in \eqref{attenuation} and \eqref{spreadloss} is the distance to the center of each bin  

\begin{equation}\label{distance_center}
    d_n^c = d_n - \frac{d_b}{2}
\end{equation}
where $d_b$ is the length of each bin.

To include the sonar resolution \eqref{resolution}, we replace the distance bins $d_n$ with resolution bins $\delta_i$. The length of each resolution bin is $\delta_y$. The total contribution from bottom backscattering for each distance bin is


\begin{equation*}
        RL_B^n = 10\log_{10}\left(\sum_{i=1}^m 10^{RL_B^i/10}\right) 
\end{equation*}
where $m$ is the largest integer such that $m\delta_y \leq (d_{n}- d_{n-1})$, $RL_B^n$ is the backscattered sound energy from the bottom for each distance bin $n$, $RL_B^i = RL_B(d_{n-1}+i\delta_y)$ is the backscattered sound energy from the bottom for each resolution bin $i$, and $d_0 = 0$.

\subsection{Surface Backscatter Area Calculation}\label{s11}
Surface backscattering contributions can be determined if the sonar depth is known. Equations \eqref{sb1} and \eqref{sb2} are used to calculate the total contribution from surface backscatter. The calculations for finding the ensonified area and beam pattern for bottom backscattering from Section \ref{s10} are used for surface backscattering with the modification that $h$ in Section \ref{s10} is the depth reading $-h_d$ (m).

\subsection{Volume Backscatter Volume Calculation}\label{s12}
The total contribution from volume backscatter is calculated using \eqref{vb1} and \eqref{vb2}. The ensonified shape of the volumes for each distance bin is a hollow sphere, minus the volume cut off by the bottom ($V_{B_{HEM}}$) and surface ($V_{S_{HEM}}$). The volumes cut off by the bottom and surface are both in the shape of hemispheres. The volume for each distance bin $n$ is 
\begin{equation*}
    V_n = \left(\frac{4}{3}\pi d_n^3 - \frac{4}{3}\pi d_{n-1}^3\right) - V_{B_{HEM}} - V_{S_{HEM}}
\end{equation*}
where 

\begin{equation*}
\begin{split}
            &V_{B_{HEM}} = \\
    &\begin{cases}
    \frac{2}{3}\pi \left( \sqrt{d_n^2-h^2} \right)^3 - \frac{2}{3}\pi \left( \sqrt{d_{n-1}^2-h^2} \right)^3
    & h>d_{n} \\
    \frac{2}{3}\pi \left( \sqrt{d_n^2-h^2} \right)^3
    & d_{n}\geq h > d_{n-1} \\
    0 & \text{otherwise}
    \end{cases}
\end{split}
\end{equation*}
and

\begin{equation*}
    \begin{split}
            &V_{S_{HEM}} =
     \\ &\begin{cases}
    \frac{2}{3}\pi \left( \sqrt{d_n^2-h_d^2} \right)^3- \frac{2}{3}\pi \left( \sqrt{d_{n-1}^2-h_d^2} \right)^3
    & h_d>d_{n} \\
    \frac{2}{3}\pi \left( \sqrt{d_n^2-h_d^2} \right)^3
    & d_{n}\geq h_d > d_{n-1} \\
    0 & \text{otherwise}
    \end{cases}
    \end{split}
\end{equation*}

The distance to the end of distance bin $n$ is $d_n$, $d_0 = 0$, the altitude is $h$ (m) and the depth is $h_d$ (m). For the method described in the remainder of this section, we do not incorporate $V_{B_{HEM}}$ or $V_{S_{HEM}}$ as this is incorporated in the beam pattern calculations instead. 

Depending on the angle the signal is transmitted and received, the signal loss from the beam pattern will vary. The horizontal beam pattern angle $\theta$, and the vertical beam pattern angle $\psi$ are required to calculate the beam pattern loss from different angles. A coordinate system is attached to the sonar transducer element such that the vector $\begin{bmatrix}1 & 0 & 0\end{bmatrix}^{\mathsf{T}}$, expressed in the coordinate frame, points directly away from the face of the transducer element. The vectors around the sphere are 

\begin{equation*}
    v_v= \begin{bmatrix} 
    \cos(\theta_h)\cos(\theta_v) \\ \sin(\theta_h)\\ 
    \sin(\theta_v)
    \end{bmatrix}
\end{equation*}
where values of $-\pi \leq \theta_h \leq \pi$ and $ -\pi \leq \theta_v \leq \pi$ are  selected uniformly around the sphere.

We use a rotation matrix to rotate each beam vector to the sonar frame. For example, a sonar directed $\theta_p$ degrees downwards from $\begin{bmatrix}1 & 0 & 0\end{bmatrix}^\mathsf{T}$, the rotation matrix times the beam vector is

\begin{equation*}
    v = \begin{bmatrix}
    \cos(\theta_p) & 0 & \sin(\theta_p) 
    \\ 0 & 1 & 0 
    \\-\sin(\theta_p) & 0 & \cos(\theta_p)
    \end{bmatrix}
    \begin{bmatrix} 
    \cos(\theta_h)\cos(\theta_v) \\ \sin(\theta_h)\\ 
    \sin(\theta_v)
    \end{bmatrix}
\end{equation*}
where the resulting $v = \begin{bmatrix}v_x & v_y & v_z\end{bmatrix}^\mathsf{T}$. The horizontal beam angle $\theta$ is
\begin{equation*}
    \theta =  \tan^{-1} \left(\frac{v_y}{v_x}\right)
\end{equation*}
The vertical beam pattern angle $\psi$ is 
\begin{equation*}
    \psi =  \tan^{-1} \left(\frac{v_z}{v_x}\right)
\end{equation*}
The average loss from the beam pattern over the entire sphere is 
\begin{equation*}
    \frac{1}{\pi^2}\int_{-\pi}^{\pi}\int_{-\pi}^{\pi}BP(\theta, \psi )d\theta_h d\theta_v
\end{equation*}
 where the beam pattern is defined 
\begin{equation*}
    BP(\theta, \psi) = 
    \begin{cases}
    BP_2(\theta, \psi) & -\frac{\pi}{2}, -\frac{\pi}{2} < \theta, \psi < \frac{\pi}{2}, \frac{\pi}{2}\\
    0 & \text{otherwise}
    \end{cases}
\end{equation*}
The angles for which the sound waves impact the bottom or surface at distance bin $n$ are
\begin{equation*}
    \theta_{ha} = \begin{cases}
        \sin^{-1}\left(\frac{2h}{d_{n}+d_{n-1}}\right) & h < \frac{d_{n}+d_{n-1}}{2} \\
        0 & \text{otherwise}    
    \end{cases}
\end{equation*}
and
\begin{equation*}
    \theta_{hd} = \begin{cases}
        \sin^{-1}\left(\frac{2h_d}{d_{n}+d_{n-1}}\right) & h_d < \frac{d_{n}+d_{n-1}}{2}  \\ 
        0 & \text{otherwise}
    \end{cases}
\end{equation*}
$BP_2$ accounts for ground and sea surface, therefore no volume reverberations will be received from beyond these angles 
\begin{equation}\label{bp_vb}
    BP_2(\theta, \psi) = 
    \begin{cases}
    20\log_{10}(\alpha \beta) \text{ dB} & - \theta_{ha} + \theta_p < \psi <  \theta_{hd}  + \theta_p \\
    0 & \text{otherwise}
    \end{cases}
\end{equation}

To compute the contribution from volume backscattering \eqref{vb1} for each distance bin, $BP_T$ and $BP_R$ are computed with \eqref{bp_vb} and the ensonified volume $V$ \eqref{vb2} is $\frac{4}{3}\pi d_n^3 - \frac{4}{3}\pi d_{n-1}^3$. The distance $d$ in \eqref{attenuation} and \eqref{spreadloss} is the distance to the center of each bin $d_n^c$ \eqref{distance_center}.

To include the sonar resolution \eqref{resolution}, we replace the distance bins $d_n$ with resolution bins $\delta_i$. The length of each resolution bin is $\delta_y$. The contribution from volume backscattering for each distance bin is


\begin{equation*}
        RL_V^n = 10\log_{10}\left(\sum_{i=1}^m 10^{RL_V^i/10}\right) 
\end{equation*}
where $m$ is the largest integer such that $m\delta_y \leq (d_{n}- d_{n-1})$, $RL_V^n$ is the backscattered sound energy from the volume for each distance bin $n$, $RL_V^i = RL_V(d_{n-1}+i\delta_y)$ is the backscattered sound energy from the volume for each resolution bin $i$, and $d_0 = 0$.

\section{Simulator}\label{SIM}

\begin{figure}[h]
\centering 
\includegraphics[width=3.0in]{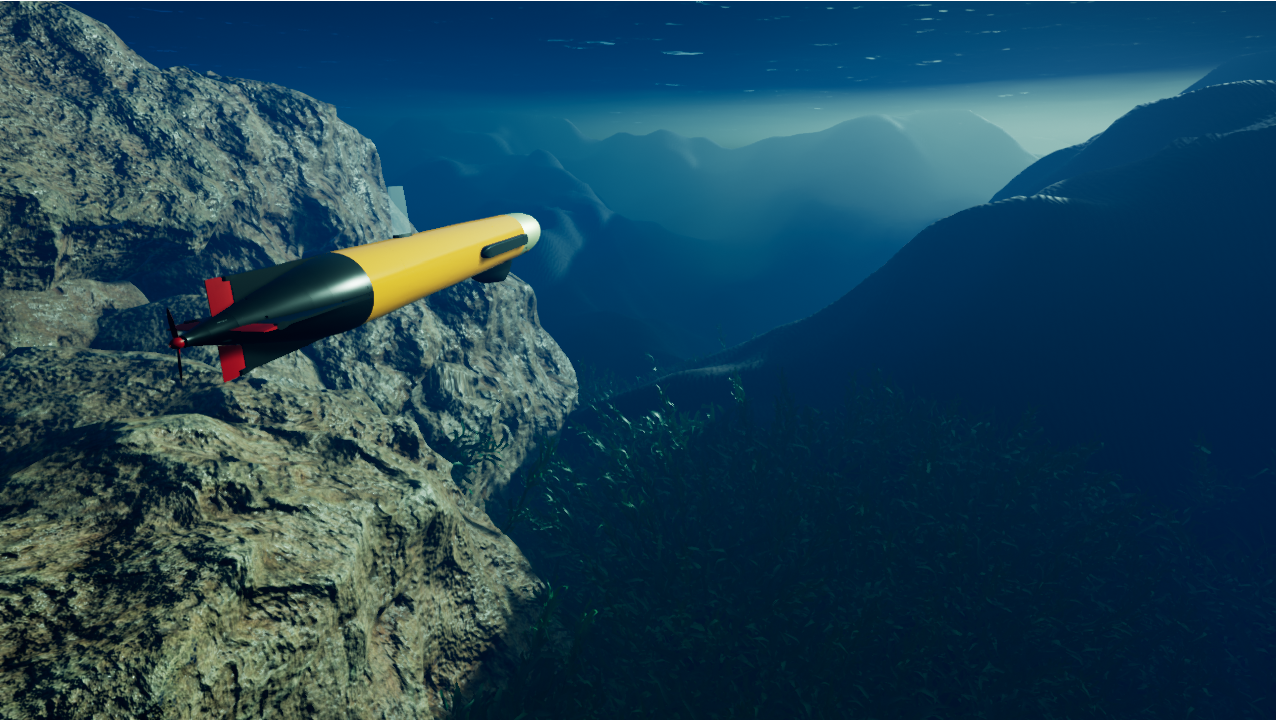}
\caption{Unreal Engine Simulation}
\label{fig1}
\end{figure}

The simulator is constructed to implement the high-fidelity sensor model. Ray tracing up to first-order multipath is employed to simulate the sonar propagation. Multiple rays are traced for each beam at the desired ping rate of the sonar. Each ray accounts for an area ahead of the sonar. If too few rays are traced, individual rays may return intensities which are too high for an object or rays can miss an object entirely. When many rays are traced, the accuracy increases, however, the computational requirements increase as well. Rays are uniformly randomly distributed around a sphere such that after multiple pings, a ray is likely to eventually hit the object. A uniform distribution of rays is achieved by sampling the $x$, $y$, and $z$ components from a Gaussian distribution. The rays are traced for the maximum sonar range $d_{\max}$ (m), determined by the ping rate $f_p$ (Hz) of the sonar,

\begin{equation}\label{d_max}
    d_{\max} = \frac{c}{2f_p}
\end{equation}

If a ray impacts either an object, the surface or the ground, the intensity of energy reflected back to the sonar is computed. If the ray impacts the surface or the bottom, the reflected energy, $I_R$, is given by $RL_B$ or $RL_S$ respectively. If the ray impacts an object, the reflected energy is

\begin{equation}\label{IR}
    I_{R} = SL - TL + BP_T + BP_R  + TS  \text{ dB}
\end{equation}
where $TS$ is the target strength of the object. The target strength is dependant on the total surface area attributed to the ray, the grazing angle of the ray, the frequency of the sonar, and the object's RMS roughness. The SEARAY model is used to compute the intensity of energy reflected back to the sonar. $I_R$ is added to the total energy received from the distance bin for which the impact occurred. Additionally, the volume reverberation received by the sonar transducer element, $RL_V$, is computed for each distance bin up to the impact and added to the corresponding bins.

\begin{algorithm}[H]
\caption{Computing Received Intensity of a Ping}
\begin{algorithmic}
\STATE \textit{// Initialize bins}
\FOR{b = 1:B} 
\STATE $I_R^b = 0$
\ENDFOR
\STATE \textit{// Compute received intensity from each ray}
\FOR {$n = 1:N$}
\STATE Trace ray $n$ for a distance of $d_{\max}$ \eqref{d_max}
\IF {impact}
\STATE Compute $I_R$ \eqref{IR}
\STATE Add $I_R$ to distance bin of impact 
\STATE \textit{// Compute $RL_V$ \eqref{vb1} for each bin up to impact bin $b$}
\FOR{m = 1 : b}
\STATE Add $RL_V$ to distance bin $m$
\ENDFOR
\STATE Compute first order multipath contribution
\ELSE 
\STATE \textit{// Compute $RL_V$ for each distance bin}
\FOR{m = 1 : B}
\STATE Add $RL_V$ to distance bin $m$
\ENDFOR
\ENDIF
\ENDFOR
\end{algorithmic}\label{alg}
\end{algorithm}

The first order multipath is computed if a ray impacts an object in the simulation environment. The ray is traced at the angle of reflection, which is the angle where highest intensity of sound energy is reflected. The multipath ray is traced for the remaining distance, up to $d_{\max}$. 

Algorithm \ref{alg} shows the pseudo code for computing the return data from a single sonar ping. 

Once the total received intensities from a sonar ping are computed, the noise level over the frequency band of the FLS is simulated. The noise level incorporates turbulence, shipping traffic, sea state and thermal noise. The number of rays traced for each ping is chosen based on the desired speed and accuracy of the simulation. 

Our simulation uses Unreal Engine 4 to visualize the environment and AUV missions, shown in Figure \ref{fig1}. The Unreal Engine portion of the simulator implements the high-fidelity sonar model and the environments in which simulations are carried out. The vehicle maneuvering model used on the VT 690 AUV \cite{c30}, built using the Robot Operating System (ROS), is integrated in Unreal Engine.

\section{Results}\label{RES}

\begin{figure}
\centering
\includegraphics[width=3.5in]{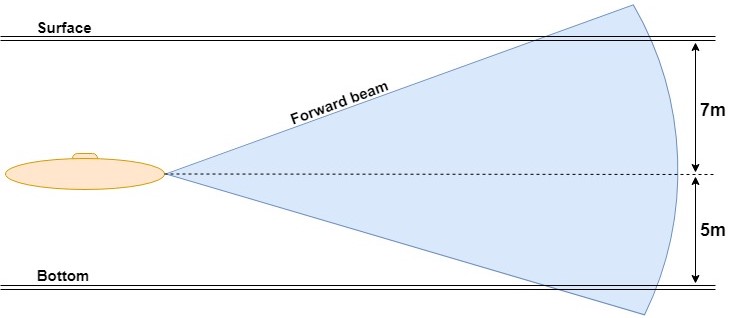}
\caption{Simulation to compare simulated vs. theoretical returns}
\label{sim1}
\end{figure}

\begin{figure}
\centering
\includegraphics[width=3.5in]{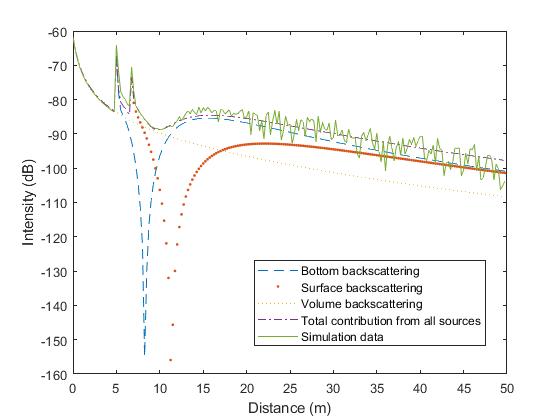}
\caption{Returns from simulation intensity vs. theoretical intensity}
\label{fig3}
\end{figure}

\begin{figure}
\centering
\includegraphics[width=3.5in]{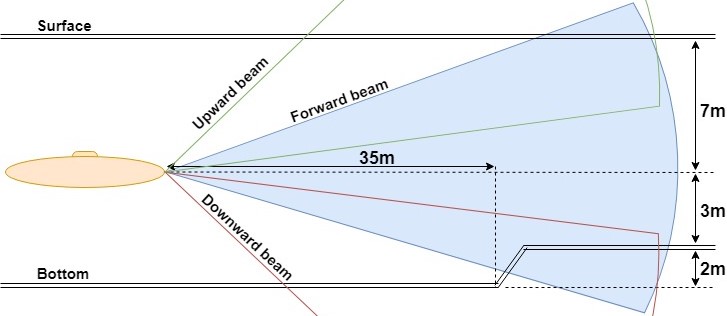}
\caption{Rise in sea floor simulation}
\label{sim2}
\end{figure}

Two scenarios are evaluated to measure the effectiveness of the simulation. The first, illustrated in Figure \ref{sim1}, is a simulation with no object in front of the AUV. The data produced by the simulator is compared to the expected returns \eqref{expected_returns}. The expected returns are computed using the process described in Sections \ref{s10}, \ref{s11}, and \ref{s12}. The second scenario, illustrated in Figure \ref{sim2}, compares the returns from three sonar beams with a rise in sea floor. For both cases, the simulation traces $20,000$ rays, the source level, $SL$, is set to $0$ dB, altitude is $5$m, depth is $7$m, bottom type is sand, wind speed is $10$ knots and particle density is low ($-90$ dB). The noise level is disabled for these two scenarios such that the simulation results can be evaluated without the noise floor.

The results for the first scenario are shown in Figure \ref{fig3} where the expected contributions from the volume, surface and bottom are shown along with the total expected returns. The expected returns are compared to the simulated data for a single forward beam (see Figure \ref{sim1}). The simulation data agrees with the theoretical returns in the closer distance bins. As the distance increases, the probability of a ray hitting the bottom or surface decreases, leading to too many or too few rays contributing to the returns from the bottom. More rays are needed in order to smooth the returns at far distances. 

\begin{figure}
\centering
\includegraphics[width=3.5in]{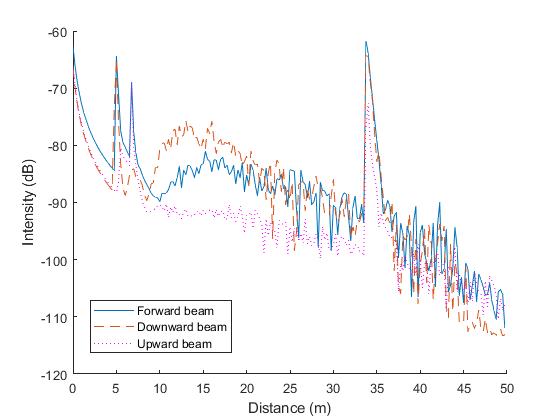}
\caption{Return from a rise in sea floor for multiple beams}
\label{fig4}
\end{figure}

Figure \ref{fig4} shows the simulation result from the second scenario, where there is an increase in sea floor height of $2$m occurring $35$m ahead of the AUV. The returns for three beams are shown, one forward beam, another angled $20$ degrees downwards and the third angled $20$ degrees upwards. The transmitter is directed forward, leading to an initial beam pattern loss when comparing the upward and downward facing beams to the forward facing beam. The first spike at $5$m shows the return from the bottom, with the downward and forward facing beams returning significantly higher than the upward beam. The second spike at 7m is from the surface, where the upward and forward beam returns are greater than that of the downward facing beam. The spike at $35$m is the object on the sea floor, which is far above the expected returns for each beam. 

The simulator is run on an Intel Core i7-6700 processor with 16 GB of RAM and an AMD Radeon RX 480 graphics card. The simulator is well suited for a limited number of beams and performs well with up to six beams each tracing $20,000$ rays. The number of beams can be scaled with an increased computational cost. When a small number of rays are traced, this leads to lower agreement with theoretical data.

\section{Conclusion}

Our analysis shows that when a sufficiently large number of rays are traced, our simulated intensity agrees with our expected performance from our sonar model. Our numerical approximations using ray tracing and environmental model bridges the gap between having a high-fidelity sonar model and AUV simulation. The proposed solution is useful for the development of an obstacle detection sonar, providing an environment to test possible configurations before manufacturing.

\bibliographystyle{IEEEtran}
\bibliography{IEEEabrv,bibi}

\end{document}